\documentclass[sigconf]{acmart}
\pdfoutput=1

\usepackage{url}
\usepackage{subfig}
\usepackage{balance}
\def\BibTeX{{\rm B\kern-.05em{\sc i\kern-.025em b}\kern-.08emT\kern-.1667em\lower.7ex\hbox{E}\kern-.125emX}}
    
\copyrightyear{2020}
\acmYear{2020}
\acmConference[EWSN 2020]{International Conference on Embedded Wireless Systems and Networks}{February 17--19, 2020}{Lyon, France}
\acmBooktitle{International Conference on Embedded Wireless Systems and Networks (EWSN 2020),
February 17--19, 2020, Lyon, France}
\acmPrice{15.00}
\acmDOI{xx.xxx/xxx.xx}
\acmISBN{xxx-xx-xx-xx-x/x/x}

\begin{document}

%
% The "title" command has an optional parameter, allowing the author to define a "short title" to be used in page headers.
\title[Closed-Loop Control over Wireless]{Demo: Closed-Loop Control over Wireless -- Remotely Balancing an Inverted Pendulum on Wheels}
%\shorttitle{\textsf{GALLOP}: Toward High-Performance Connectivity for closing Control Loops}
%
% The "author" command and its associated commands are used to define the authors and their affiliations.
% Of note is the shared affiliation of the first two authors, and the "authornote" and "authornotemark" commands
% used to denote shared contribution to the research.

\author{Aleksandar Stanoev, Adnan Aijaz, Anthony Portelli, and Michael Baddeley  }
\affiliation{%
 \institution{Bristol Research and Innovation Laboratory\\ Toshiba Research Europe Ltd., Bristol, U.K.}}
 %\streetaddress{32 Queen Square}
 %\city{Bristol}
 %\country{U.K.}}
 \email{adnan.aijaz@toshiba-trel.com}
 %\email{aleksandar.stanoev@toshiba-trel.com}

%\author{Adnan Aijaz}
%\affiliation{%
% \institution{Toshiba Research Europe Ltd.}
% \streetaddress{32 Queen Square}
% \city{Bristol}
% \country{U.K.}}
% \email{adnan.aijaz@toshiba-trel.com}
% 
%
%
%\author{Anthony Portelli}
%\affiliation{%
% \institution{Toshiba Research Europe Ltd.}
% \streetaddress{32 Queen Square}
% \city{Bristol}
% \country{U.K.}}
% \email{anthony.portelli@toshiba-trel.com}
%
%<Michael.Baddeley@toshiba-trel.com>

%
% By default, the full list of authors will be used in the page headers. Often, this list is too long, and will overlap
% other information printed in the page headers. This command allows the author to define a more concise list
% of authors' names for this purpose.
\renewcommand{\shortauthors}{Stanoev et al.}

%
% The abstract is a short summary of the work to be presented in the article.
\begin{abstract}
Achieving closed-loop control over wireless is crucial in realizing the vision of Industry 4.0 and beyond. This demonstration shows the viability of closed-loop control over wireless through a high-performance wireless solution.  The closed-loop control problem involves remote balancing of a two-wheeled robot that represents an inverted pendulum on wheels.

%Various legacy and emerging industrial applications demand closed-loop control over multiple hops. Existing solutions
%for control-aware wireless design are mostly focused on singlehop
%connectivity. Moreover, state-of-the-art multi-hop wireless
%technologies do not fulfill the performance requirements of
%closed-loop control. This paper proposes a novel wireless solution,
%termed as \textsf{GALLOP}, for realizing closed-loop control over multihop
%networks, with fast dynamics on the order of few milliseconds
%(ms). \textsf{GALLOP} is based on a pragmatic design approach that
%specifically accounts for the peculiarities of closed-loop control.
%Key aspects of \textsf{GALLOP} design include control-aware bidirectional
%multi-hop scheduling for cyclic information exchange,
%efficient signaling mechanism for schedule construction, and
%robust retransmission techniques based on cooperative multiuser
%diversity. Performance evaluation based on extensive systemlevel
%simulations and hardware implementation on a Bluetooth
%5 testbed demonstrates the viability of \textsf{GALLOP} in providing
%high-performance connectivity with very low and deterministic
%latency, very high reliability and high scalability, to meet the
%stringent requirements of closed-loop control over multi-hop
%wireless networks.
\end{abstract}

%\begin{CCSXML}
%<ccs2012>
%  <concept_id>10010520.10010553.10010562</concept_id>
%  <concept_desc>Computer systems organization~Embedded systems</concept_desc>
%  <concept_significance>500</concept_significance>
%  <concept_id>10010520.10010575.10010755</concept_id>
%  <concept_desc>Computer systems organization~Redundancy</concept_desc>
%  <concept_significance>300</concept_significance>
%  <concept_id>10010520.10010553.10010554</concept_id>
%  <concept_desc>Computer systems organization~Robotics</concept_desc>
%  <concept_significance>100</concept_significance>
%  <concept_id>10003033.10003083.10003095</concept_id>
%  <concept_desc>Networks~Network reliability</concept_desc>
%  <concept_significance>100</concept_significance>
%</ccs2012>
%\end{CCSXML}
%
%\ccsdesc[500]{Computer systems organization~Embedded systems}

%
% Keywords. The author(s) should pick words that accurately describe the work being
% presented. Separate the keywords with commas.

%
% A "teaser" image appears between the author and affiliation information and the body 
% of the document, and typically spans the page. 
%  \Description{Enjoying the baseball game from the third-base seats. Ichiro Suzuki preparing to bat.}

%
% This command processes the author and affiliation and title information and builds
% the first part of the formatted document.
\maketitle

\section{Introduction}
%ACM's consolidated article template, introduced in 2017 \cite{mh_cont}, provides a consistent \LaTeX\ style for use across ACM publications, and incorporates accessibility and metadata-extraction functionality necessary for future Digital Library endeavors. Numerous ACM and SIG-specific \LaTeX\ templates have been examined, and their unique features incorporated into this single new template.

%Recent advances in wireless technologies like WirelessHART \cite{wirelesshart}, ZigBee \cite{zbee} and ISA100.11a \cite{isa} have led to the use of multi-hop wireless networks for (open-loop) monitoring of large-scale industrial systems. However, the use of multi-hop wireless networks for control
%applications is still at a nascent stage. 

Closed-loop control or feedback control is one of the most prominent industrial control applications \cite{TI_PIEEE}. Typically, closed-loop control takes place between a controller and a spatially distributed system of sensors and actuators. Closed-loop control represents a networked control system wherein control and information packets are exchanged over a shared medium thereby closing a global control loop. 

Closed-loop control poses significant challenges to the underlying communication network. First, it demands real-time connectivity with very low latency and very high reliability. Second, it involves bi-directional information exchange with cyclic traffic patterns which requires highly deterministic connectivity, i.e., communication latency between cycles must have very low variance. Third, the presence of a large number of sensors and actuators creates the requirement of highly scalable connectivity. 

In legacy industrial applications like factory automation, closed-loop control is realized through wired connectivity technologies like fieldbus systems or industrial Ethernet. However, wired connectivity incurs high installation and maintenance costs. Moreover, it is not suitable for many new industrial closed-loop control applications like formation control of automated guided vehicles, highway platooning and collaborative remote operation which are characterized by the requirements of mobility and flexibility. 

%\textsf{GALLOP} \cite{gallop_conf}, \cite{gallop_patent}

To this end, the main objective of this demonstration is to show the viability of real-time closed-loop control over wireless. We consider the case of remotely balancing a two-wheeled robot that represents an inverted pendulum on wheels. We leverage a high-performance wireless technology, known as \textsf{GALLOP} \cite{gallop_conf}, \cite{gallop_patent}, developed in our previous works and specifically designed for meeting the stringent requirements of closed-loop control.

%\end{itemize}

\section{Demonstration Overview}
\begin{figure*}
\centering
\includegraphics[scale=0.45]{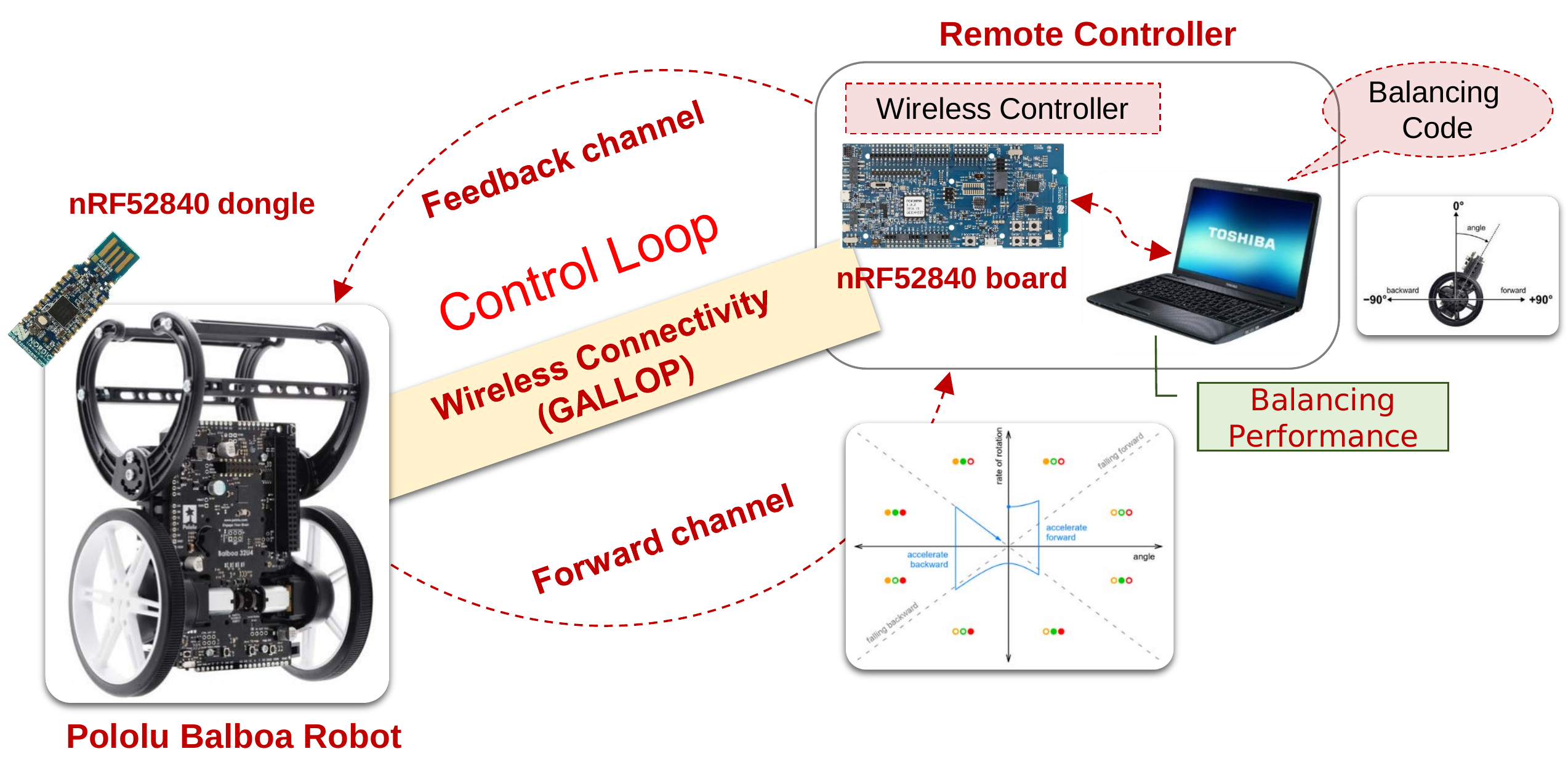}
\caption{Closed-loop control scenario and configuration. }
%\vspace{-0.4cm} 
\label{scenario}
\end{figure*}

The inverted pendulum is a classic closed-loop control problem in control theory and widely used in benchmarking control strategies. The inverted pendulum is naturally unstable without active control. A two-wheeled robot actually depicts an inverted pendulum on wheels. Conventionally, such a robot balances itself through an onboard closed-loop control algorithm based on wired connectivity between a controller and one or more sensors/actuators.  Our objective is to realize such closed-loop control over wireless. The demonstration scenario is illustrated in Fig. \ref{scenario}. We decouple part of the overall balancing functionality from the robot such that the balancing algorithm runs in a remote controller which communicates with the robot over wireless. 

\section{Design and Implementation}
\subsection{Balancing Robot}
We use the Balboa 32U4 balancing robot by Pololu\footnote{https://www.pololu.com/} that is both programmable and customizable. Its control board is built around an Atmel ATmega32U4 AVR microcontroller. The control board features two H-bridge motor drivers, quadrature encoders and a complete inertial measurement unit (IMU) comprising a 3D accelerometer, a 3D gyro and a 3-axis magnetometer that allows it to determine its orientation. The robot uses two micro metal gearmotors to drive external 2-gear gearboxes. Our robot uses 80 x 10 mm wheels.

\subsection{Balancing Algorithm}
\begin{figure}
\centering
\includegraphics[scale=0.32]{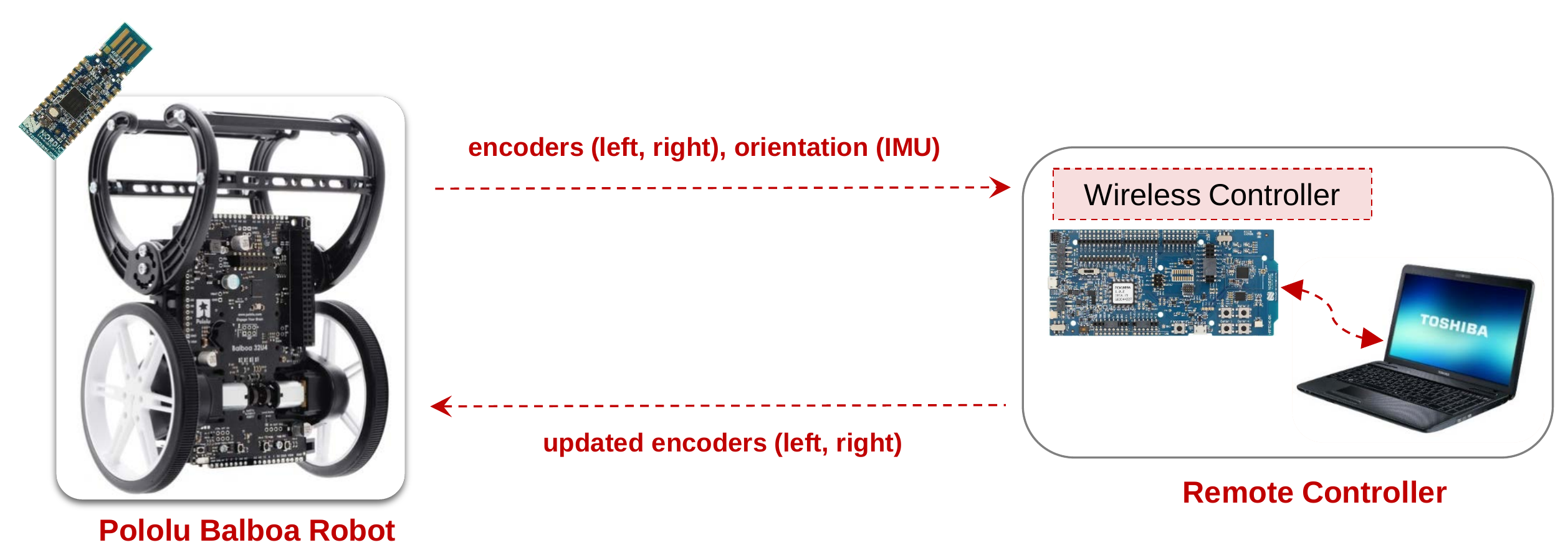}
\caption{Illustration of remote balancing operation. }
%\vspace{-0.4cm} 
\label{param}
\end{figure}

The balancing functionality is achieved through a proportional-integral-derivative (PID)-like feedback control algorithm.  The key to remotely balancing the Balboa robot is the nine sensors pertaining to the IMU that provide  a sense of its orientation in 3D. On the forward channel, the robot sends the orientation information along with the encoder values to the wireless controller. The balancing algorithm computes robot's rate of rotation and sends updated motor speed on the feedback channel. The remote balancing operation is illustrated in Fig. \ref{param}.

\subsection{Wireless Design}
The demonstration employs \textsf{GALLOP} as the underlying wireless technology for communication between the robot and the remote controller. \textsf{GALLOP} has been specifically designed to account for the peculiarities of closed-loop control. The medium access control (MAC) layer of \textsf{GALLOP} is based on time division multiple access (TDMA), frequency division duplexing (FDD) and frequency hopping.  \textsf{GALLOP} provides very low and deterministic latency which is required for closed-loop control. \textsf{GALLOP} also implements a number of techniques for achieving very high reliability. \textsf{GALLOP} implements a flooding-based protocol for time synchronization \cite{glossy}.  Our \textsf{GALLOP} implementation is based on the 2 Mbps Physical (PHY) layer of the latest Bluetooth 5.0 standard. We have implemented \textsf{GALLOP} on the Nordic nRF52840\footnote{https://www.nordicsemi.com/Products/Low-power-short-range-wireless/nRF52840} platform. The robot is equipped with an nRF52840 dongle. 

\subsection{Performance Evaluation}

\begin{figure}
\centering
\includegraphics[scale=0.29]{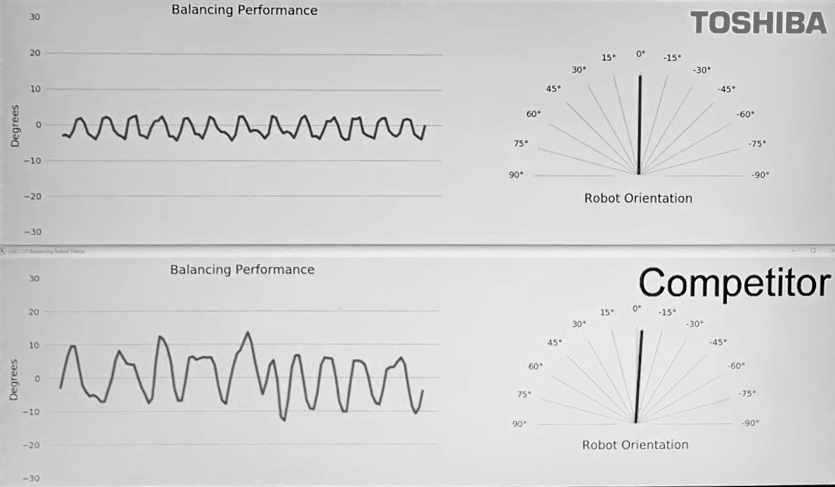}
\caption{A snapshot from the demonstration showing balancing performance. }
%\vspace{-0.4cm} 
\label{perf}
\end{figure}

We quantify the balancing performance in terms of robot's forward/backward rate of rotation in degrees. As shown in Fig. \ref{perf}, balancing performance over \textsf{GALLOP} shows a deterministic behavior with minimal rate of rotation. To achieve accurate balancing performance, the cyclic latency of the control loop should be less than 10 ms. With \textsf{GALLOP}, the communication latency is approximately 2 ms which provides sufficient margin for processing overheads and software delays. The competitor depicts a standard Bluetooth 5.0 protocol with  minimal possible communication latency of 7.5 ms. As shown in the results, this leads to  non-deterministic behavior with larger rate of rotation. A video of the demonstration is available at \url{https://www.dropbox.com/s/le1r851dl1gzpvl/Balancing_Robot.mp4?dl=0}

\section{Remarks}
Achieving closed-loop control over wireless is crucial in realizing the envisioned transformation of industrial systems. This demonstration reveals that the stability of closed-loop control over wireless is heavily dependent on deterministic as well as low latency of the underlying communication technology.  
%A short video \cite{demo_infocom} of \textsf{GALLOP} demonstration  for real-time closed-loop control pertaining to remote control as well as formation control of mobile platforms is available at \url{https://www.dropbox.com/s/roctb3pac5o8y53/GALLOP_demo_final.mp4?dl=0}
%\section{SIGCHI Extended Abstracts}
%
%The ``\verb|sigchi-a|'' template style (available only in \LaTeX\ and not in Word) produces a landscape-orientation formatted article, with a wide left margin. Three environments are available for use with the ``\verb|sigchi-a|'' template style, and produce formatted output in the margin:
%\begin{itemize}
%\item {\verb|sidebar|}:  Place formatted text in the margin.
%\item {\verb|marginfigure|}: Place a figure in the margin.
%\item {\verb|margintable|}: Place a table in the margin.
%\end{itemize}

%
% The acknowledgments section is defined using the "acks" environment (and NOT an unnumbered section). This ensures
% the proper identification of the section in the article metadata, and the consistent spelling of the heading.
%\begin{acks}
%To Robert, for the bagels and explaining CMYK and color spaces.
%\end{acks}
\begin{figure*}
\centering
\includegraphics[scale=0.4]{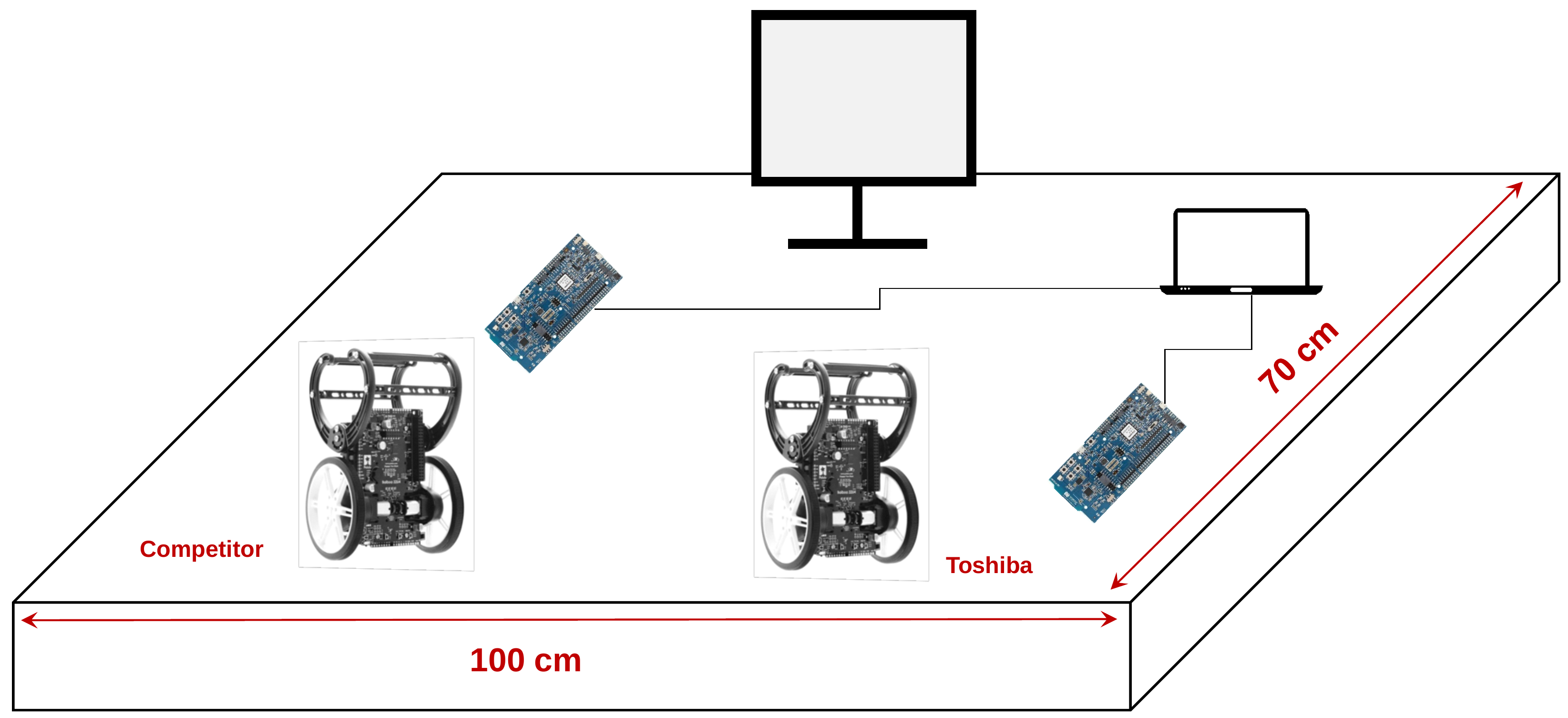}
\caption{Layout of the demonstration.}
%\vspace{-0.4cm} 
\label{reqs}
\end{figure*}

% The next two lines define the bibliography style to be used, and the bibliography file.
\balance
\bibliographystyle{ACM-Reference-Format}
%\bibliography{sample-base}
\bibliography{GALLOP_bib}
%\printbibliography{\textsf{\textsf{GALLOP}}_bib.bib}
% If your work has an appendix, this is the place to put it.
\appendix
\newpage

\section*{Appendix}
The demonstration requirements are highlighted in Fig. \ref{reqs}. A desk with approximate dimensions of 100 cm x 70 cm is required to display the following hardware components. 

\begin{itemize}
\item 2  Pololu Balboa 32U4 balancing robots. 
\item 2 Nordic nRF52840 wireless modules. 
\item 1 monitor/display (to be provided by demo chairs).
\item 1 laptop. 
\end{itemize}

\end{document}